\documentclass[runningheads]{llncs}
\usepackage{graphicx}
\usepackage{amsmath,amssymb} 
\usepackage{color}
\usepackage{gensymb}

\begin{document}
\title{Connecting Gaze, Scene, and Attention: Generalized Attention Estimation via Joint Modeling of Gaze and Scene Saliency}

\titlerunning{Connecting Gaze, Scene, and Attention}
%
\author{Eunji Chong \and Nataniel Ruiz \and Yongxin Wang \and Yun Zhang \and Agata Rozga \and James M. Rehg}
%
\authorrunning{E. Chong, N. Ruiz, Y. Wang, Y. Zhang, A. Rozga, and J. M. Rehg}
%

\institute{School of Interactive Computing, Georgia Institute of Technology, Atlanta, GA, USA\\
\email{\{eunjichong,nataniel.ruiz,ywang751,yzhang467,agata,rehg\}@gatech.edu}}
\maketitle              
\begin{abstract}
This paper addresses the challenging problem of estimating the general visual attention of people in images. Our proposed method is designed to work across multiple naturalistic social scenarios and provides a full picture of the subject's attention and gaze. In contrast, earlier works on gaze and attention estimation have focused on constrained problems in more specific contexts. In particular, our model explicitly represents the gaze direction and handles out-of-frame gaze targets. We leverage three different datasets using a multi-task learning approach. We evaluate our method on widely used benchmarks for single-tasks such as gaze angle estimation and attention-within-an-image, as well as on the new challenging task of generalized visual attention prediction. 
In addition, we have created extended annotations for the MMDB and GazeFollow datasets which are used in our experiments, which we will publicly release.

\end{abstract}
%
%
%

\section{Introduction}
As humans, we are exquisitely sensitive to the gaze of others. We can rapidly infer if another person is making eye contact, follow their gaze to identify their gaze target, categorize quick glances to objects, and even identify when someone is not paying attention~\cite{land2009looking}. Automatically detecting and quantifying these types of visual attention from images and video remains a complex, open challenge. Although gaze estimation has long been an active area of research, most work has focused on relatively constrained versions of the problem in specific predetermined contexts. For example, \cite{zhang15_cvpr,cvpr2016_gazecapture} predict the gaze target \textit{given} that the person is looking at a point on a smartphone screen, \cite{nips15_recasens} predicts fixation on an object \textit{given} that the person is looking at salient object within the frame, \cite{chong2017detecting,zhang2017everyday} predict eye contact \textit{given} that the camera is located near the subject's eyes, and \cite{Recasens_2017_ICCV} predicts the focus of a person's gaze across views in commercial movies which include camera views that follow the actor's attention. It remains a significant challenge to design a system that can model the visual attention of subjects in unconstrained scenarios, without the preconditions utilized by prior works. We call this the problem of~\textit{generalized visual attention prediction}.

\begin{figure*}[t]\label{gaze_intro_example}
\centering
  \includegraphics[width=0.9\textwidth]{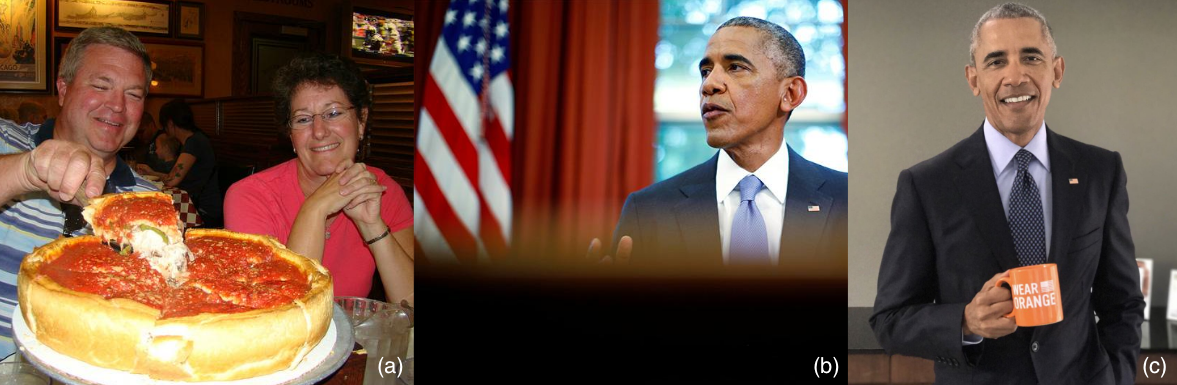}
  \caption{We present a model which aims to understand different aspects of the generalized attention prediction problem which are exemplified above. In (a) subjects are looking at a salient object in the scene, in (b) the subject is looking somewhere outside of the frame and in (c) the subject is looking at or around the camera. Our model predicts the 3D gaze vector of subjects in each of these images, along with the location of the gaze fixation in the image, if it exists. Our model explicitly determines if the subject's gaze target lies outside the frame.}
\end{figure*}

The three examples in Figure \ref{gaze_intro_example} illustrate the difficulty of the challenge. In Figure~\ref{gaze_intro_example} (a), the subjects are looking at a salient object in the scene, while in (b) the subject is looking somewhere outside of the scene, and (c) they are looking at the camera. The case of Figure~\ref{gaze_intro_example} (a) is addressed by the pioneering work of Recasens et al.~\cite{nips15_recasens}, which tackles this problem by obtaining human annotations of subject gaze targets, leveraging the finding from \cite{li2014secrets,borji2015salient,borji2013stands} which indicate that annotators very often agree on which object is salient in the scene. 
Their approach, however, was not designed to handle cases (b) and (c) since the dataset annotation process forces human annotators to label a point in the image as the fixation location. In other words the dataset does not distinguish between subjects looking at a point inside of the image or looking somewhere outside of the image. A purely saliency based approach would also fail: notice that there are salient objects in (b), an American flag, and (c), a mug, which can confound such an approach.

Figure~\ref{gaze_intro_example} (c) corresponds to the case of screen-based eye tracking~\cite{cvpr2016_gazecapture,FunesMora_ETRA_2014}, in which subjects look at images on a screen and are captured by a calibrated camera that permits the estimation of the gaze location. 
The scenario in Figure~\ref{gaze_intro_example} (a) corresponds to gaze following and has been addressed in~\cite{nips15_recasens}. 
Figure~\ref{gaze_intro_example} (b) represents a challenging case, which has not been addressed by prior work, in which the gaze target lies outside the frame and therefore cannot be identified explicitly without additional information.

It is challenging to design an attention model which can deal with these three different scenarios reliably. We tackle this problem by developing a novel generalized visual attention estimation method which jointly learns a subject-dependent saliency map and a 3D gaze vector represented by yaw and pitch. This allows us to estimate the final fixation likelihood map.

Our method is designed so that the fixation likelihood map becomes close to zero when the subject is looking outside the frame, as in cases (b) and (c). When the subject is looking at a target which is visible in the image, as in case (a), then the fixation likelihood map predicts where the subject is likely to be attending. The model simultaneously estimates the 3D gaze angle to provide a complete picture of the subject's attention and gaze. As a result, our approach produces interpretable results spanning all three of the cases in Figure~\ref{gaze_intro_example}. 
 
\textit{Our Contribution.} The major contribution of this work is our method for \textit{generalized visual attention prediction}, which works across most natural scenarios. To effectively train our model, we exploit three public datasets that were originally collected for different tasks. Specifically, we use the EYEDIAP dataset~\cite{FunesMora_ETRA_2014} to learn a precise gaze angle representation, a modified version of the GazeFollow dataset~\cite{nips15_recasens} to learn a gaze-relevant scene saliency representation, and the SynHead dataset~\cite{Gu_2017_CVPR} to complement the first two datasets as it includes large face pose variations and subject attention outside of the image frame.

As a result of our multi-task learning approach, our model achieves state-of-the-art results on the GazeFollow~\cite{nips15_recasens} task, which consists in identifying the location of the scene the subject is looking at. Our model also competes with state-of-the-art models on the 3D gaze estimation task from the EYEDIAP dataset~\cite{FunesMora_ETRA_2014}. Most importantly, we evaluate our full model on a new challenging task that automatically quantifies dense visual attention in naturalistic social interactions. We report our results on the Multimodal Dyadic Behavior (MMDB) dataset~\cite{rehg2013decoding}, a dataset of video recordings of the social and communicative behavior of toddlers. This dataset has frame-level annotations of subject's visual targets among many other nonverbal behaviors. We are the first to report attention estimation results on this dataset. We compare our results to several baselines, demonstrating the superior performance of our method.

\section{Related work}
\textit{Gaze Estimation:} Gaze estimation aims to predict the gaze of a human subject. Our work is related to third person gaze estimation and tracking methods which seek to estimate either the three dimensional direction of the gaze or the fixation point of the gaze on a screen. Krafka et al. \cite{cvpr2016_gazecapture} predict the coordinates of the gaze on a smartphone screen and present a dataset which addresses this problem. Mora et al.~\cite{FunesMora_ETRA_2014} present EYEDIAP, a dataset designed for the evaluation of gaze estimation which was collected in a controlled lab environment. They devise an RGB-D method which predicts the 3D vector of the gaze of the subject. There exist datasets which address similar tasks, such as MPIIGaze~\cite{zhang15_cvpr}, as well as synthetic datasets of eye images for gaze estimation~\cite{wood2015rendering,sugano2014learning}.
In addition to predicting the 3D gaze vector, our work predicts a fixation likelihood map of the scene as well as whether the person is looking at a location inside or outside of the image.\\
\textit{Visual Saliency:} The objective of visual saliency prediction is to estimate locations in an image which attract the attention of humans looking \textit{at} the image. Since the seminal work of Itti et al.~\cite{itti1998model} visual saliency prediction has been extensively studied. Recently deep learning methods have shown superior performance on this task due to their ability to learn features and to incorporate both local and global context into the prediction~\cite{wang2015deep,li2015visual,zhao2015saliency}. Our work in generalized visual attention prediction is influenced by the task of visual saliency since people tend to look at salient objects inside a scene, yet it is distinct because we consider cases where the subject is not looking at any object in the scene. A method driven primarily by saliency detection would not succeed in the latter case. Furthermore, real-world scenes are more likely to generate a wide range of 3D gaze directions in comparison to a screen-based eye-tracking scenario.\\
\textit{Gaze Following:} The paper by Recasens et al.~\cite{nips15_recasens} presents a new computer vision problem which inspired this work. The problem can be described as follows: given a single image containing one or more people, predict the location that each person in the scene is looking at. They presented a novel dataset which contains manual annotations of where the subjects are looking in each image. Our work differs in that we consider cases where the subjects are looking \emph{outside} of the frame, and we predict the gaze direction in these cases even though the gaze target is not visible. In addition to predicting a fixation likelihood map for the image, we predict the 3D gaze vector of each subject. Gorji and Clark~\cite{gorji2017attentional} study a problem at the intersection of visual saliency and gaze following, which consists of incorporating signals from regions of an image which guide attention to a certain part of the image. For example, when subjects in an image look at an object, this amplifies the apparent saliency of the object. Again, our problem differs in that we do not predict visual saliency but we predict the subject's gaze fixation and gaze direction.\\
\textit{Attention Modeling:} Prior works have presented different methods for measuring third-person visual attention using an environment-mounted camera. By assuming body or head orientation is a good proxy for visual orientation,~\cite{benfold2009attention} projects attention on the street by tracking pedestrians in 3D,~\cite{soo2015social,Cristani} model the focus of attention in crowded social scene, and~\cite{chen2016subjects} predicts the object in the scene that a person is interacting with which is usually indicated by hand manipulation or pointing. Our work is certainly related, although it differs because we explicitly consider the gaze of the subject.\\

\section{Method}
\begin{figure}[ht]
  \centering
      \includegraphics[width=1\textwidth]{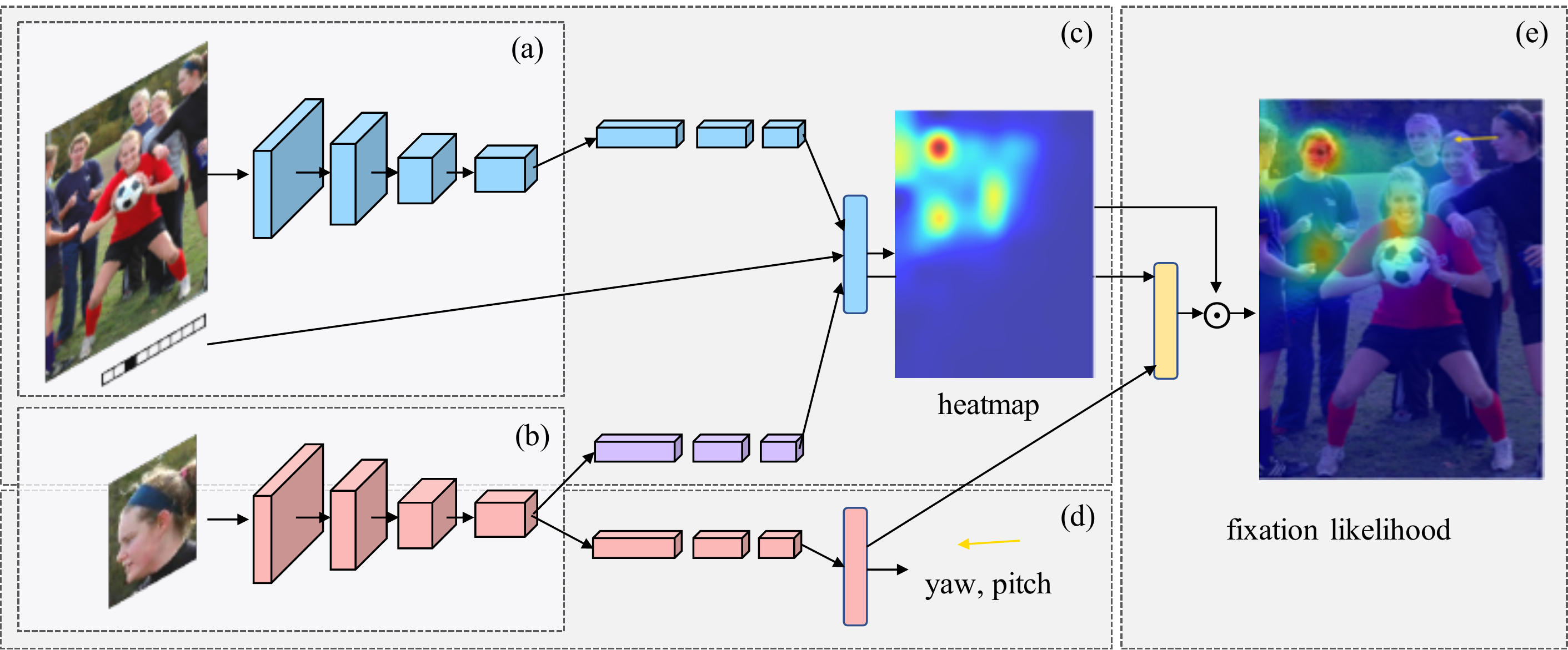}
  \caption{\textit{Overview of our approach.} Full scene image, a person's face location whose visual attention we want to predict, and the corresponding close-up face image is provided as input. Scene and face images go through separate convolutional layers in such a way that (a) (b) and (c) contribute to person-centric saliency, and (b) and (d) contribute to gaze angle prediction. In the very last layer, the final feature vectors for these two tasks are combined to estimate how likely the person is actually fixating at a gaze target within the observable scene.}\label{fig:overview}
\end{figure}

Figure~\ref{fig:overview} is an overview of our deep neural network model and its input and output. The model takes three inputs: the whole image, a crop of the subject's face, and the location of her face. Given the input, the model estimates 1. the subject's gaze angle in terms of yaw and pitch degrees (``where'' component of visual attention), 2. the subject-dependent saliency in terms of a heatmap (``what'' component of visual attention), and 3. how likely the subject is fixating at the estimated gaze target in the scene (overall ``strength'' of visual attention). 

The model has two fully-convolutional pathways, one connected to the whole image (Figure~\ref{fig:overview}-a) and the other connected to the face image (Figure~\ref{fig:overview}-b). The reasoning behind having two separate pathways is inspired by the way humans infer another person's visual attention, as first exploited by~\cite{nips15_recasens}. For example, when we interpret a person's attention from an image, we infer their gaze direction and consider whether there are any salient objects in the image along the estimated direction. Based on this hypothesis, \cite{nips15_recasens} connects two independent conv pathways together to learn the heatmap (Figure~\ref{fig:overview}-c). We take this approach further and extend their model by explicitly training for the gaze angle (Figure~\ref{fig:overview}-d) with a convolutional pathway that is connected to the face image, using a multi-task learning framework. Adding the gaze angle output as an auxiliary task has several advantages, including the additional supervisory signal that we can devise based on the relationship between gaze heatmap and angle, which pushes performance in heatmap estimation even further.

Lastly, we define the likelihood of fixation: a single-valued measure of how likely it is that the subject is looking at the estimated target region inside the frame. It is modeled by a fully connected layer (Figure~\ref{fig:overview}-e). Using this last output, the model can produce a much more complete estimation of a person's visual attention. Think of the case of Figure~\ref{gaze_intro_example}-b or c, where the person is looking outside the image frame. In such cases, we want the heatmap to be as close to zero as possible since the person is not attending to any point inside of the image. By training this last layer to produce higher value for when it is more certain that the heatmap region is attended to and lower value otherwise, the value can be applied to the heatmap with an operator $\odot$ which can be a weighting operator or a gating operator depending on application.

Since there exists no single dataset that covers all of the various gaze and scene combinations that we address in this paper (e.g., looking outside the frame, looking at the camera, fixation on an in-frame object, etc), we adopt a cross domain learning approach where the model learns partial information relevant to each task from different datasets. Depending on what supervisory signal is available in a given batch of training data, the model selectively updates its corresponding branches.

We describe the model architecture in more detail in subsection \ref{subsec:model}. We elaborate on the loss function in \ref{subsec:loss} and talk about the datasets and training procedures in \ref{subsec:crossdom}.

\subsection{Model}\label{subsec:model}
The inputs given to the model are the entire image, the subject's cropped face and the location of the face of the subject whose attention we want to estimate. The two images are resized to 227$\times$227 so that the face can be observed in higher resolution by the network. Face position is available in terms of the $(x,y)$ full image coordinates. These coordinates are quantized into a 13$\times$13 grid and then flattened to a 169 dimensional 1-hot vector.

The model consists of two convolutional (conv) pathways: a face pathway (Figure~\ref{fig:overview}-d) and a scene pathway (Figure~\ref{fig:overview}-c). 
ResNet 50~\cite{He2015} is used as a backbone network for the conv pathways (Figure~\ref{fig:overview}-a and b). Specifically we use all conv layers of ResNet50 for each of our conv pathways. After each ResNet50 block we add three conv layers (1x1, followed by 3x3, followed by 1x1) with ReLu and batch norm - with stride 1 and no padding. The blue conv layers represented in (c) have filter depth of 512, 128 and 1 respectively. The purple and red conv layers after the face pathway (represented in (c) and (d)) have filter depth of 512, 128 and 16. These conv layers serve to reduce the dimensionality of the features extracted by the ResNet50 backbone networks.

In the face pathway, the feature vector computed with the face input image goes through a fully connected layer to predict the gaze angle represented using yaw and pitch intrinsic Euler angles. In the scene pathway, the feature vectors extracted from the whole image as well as from the face image are concatenated with the face position input vector to learn the person-centric heatmap. Similarly to face position, the ground truth used for learning the heatmap is available as a gaze target position in $(x,y)$ coordinates which is quantized into 10 grids in each dimension.

Lastly, the input vectors to the last layer of each pathway are concatenated and go into the final fully connected layer to estimate the ``strength'' of the fixation ie. how likely it is that the person is actually fixating at a gaze target within the observable scene. The training label for this value is equal to 1 for a fixation inside of the image and 0 when the subject is looking outside of the scene. We also explore alternative model architectures and restrict our training to a subset of the three datasets. Experiments are reported in Section~\ref{sec:model_diagnostic}.

\subsection{Loss}\label{subsec:loss}
\begin{figure}[t]
  \centering
      \includegraphics[width=0.8\textwidth]{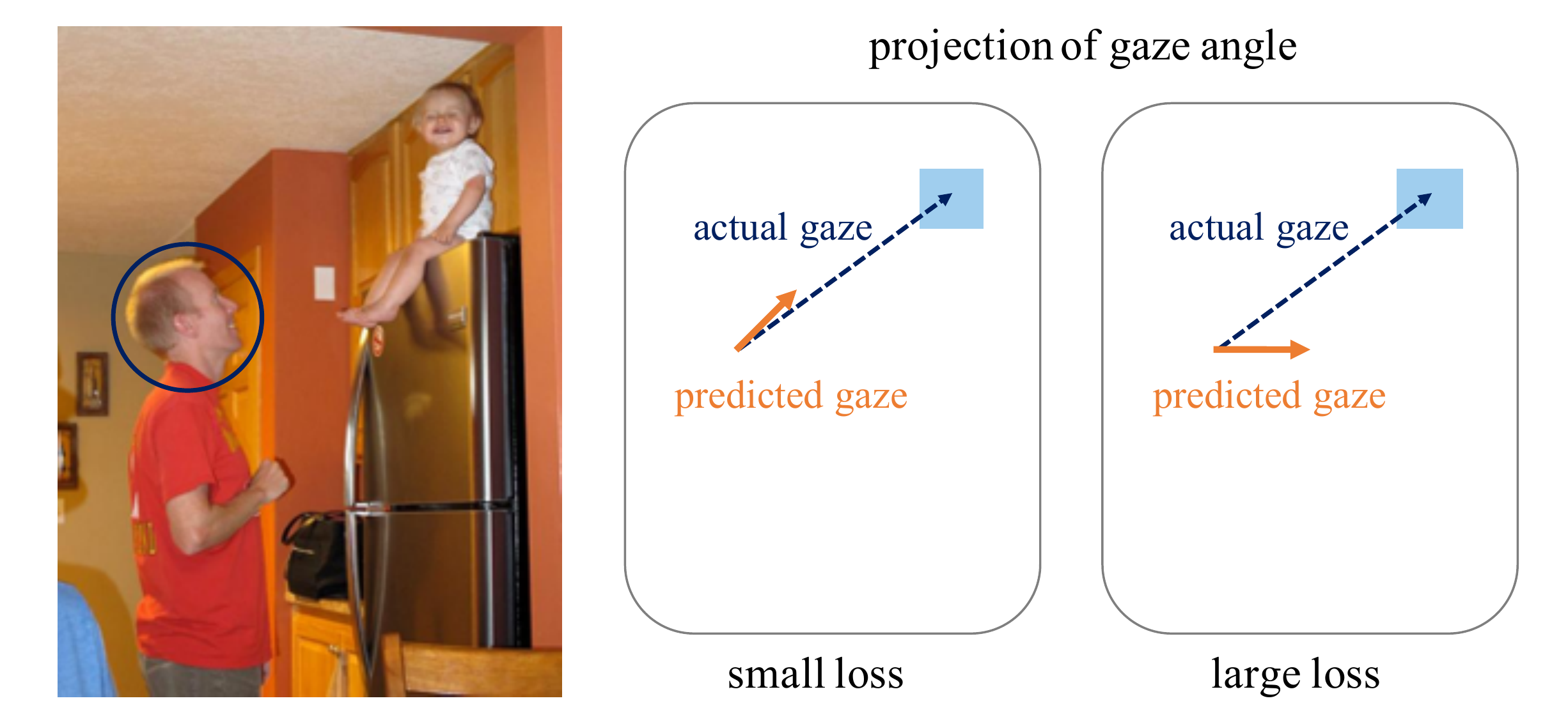}
  \caption{Our project and compare loss is illustrated here. If the estimated angle is close to the actual one, the projected gaze angle on the image should also be close to the vector connecting the head position to the gaze target.}\label{fig:pnc}
\end{figure}

As our model predicts gaze angle, saliency map and the fixation likelihood, we need to apply appropriate loss functions for each task. For the angle regression task we use an \textit{L1 loss}, and for the other two tasks we use a \textit{cross entropy loss}.
Moreover, we recognize that the gaze angle and fixation target predictions are closely related. Based on their relationship additional constraints can be imposed to augment the training loss signal. Namely, when the subject is looking at a target, the actual gaze is a ray from the subject's head to the gaze target. This ray can be projected onto the image. It becomes a 2D vector coming from the subject's head to the target exemplified by the blue vector in Figure~\ref{fig:pnc}. If the estimated angle is close to the actual one, the projected gaze angle on the image (orange vector in Figure~\ref{fig:pnc}) should also be close to the blue vector. The proximity of the two vectors is measured using the cosine distance. We call this the \textit{project and compare loss}.

\subsection{Cross-Domain Datasets and Training Procedure}\label{subsec:crossdom}
\begin{figure}[h]
  \centering
      \includegraphics[width=1\textwidth]{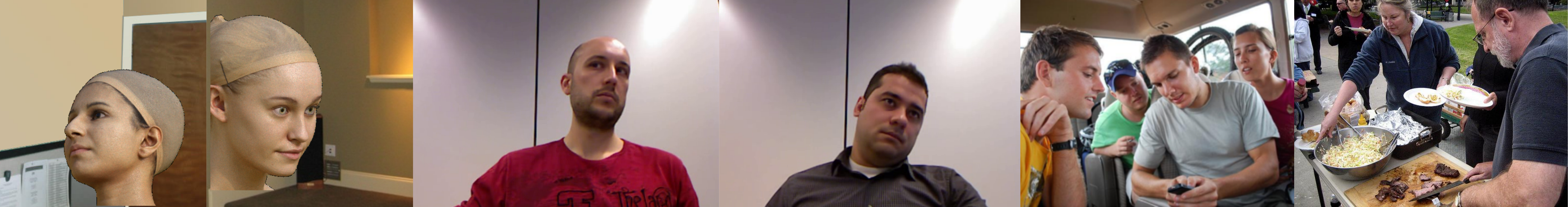}
  \caption{Examples of datasets used to train our model. Left two: SynHead, middle two: EYEDIAP, right two: GazeFollow.}\label{fig:dataset}
\end{figure}

The largest challenge in training our model is the lack of availability of training examples. Although there are a couple of existing datasets that are suitable for training certain parts of our network, no single dataset contains all of the information that we need to train the full model. Therefore, we leverage three different datasets, namely, GazeFollow~\cite{nips15_recasens}, EYEDIAP~\cite{FunesMora_ETRA_2014}, and SynHead~\cite{Gu_2017_CVPR}. We selectively train different sub-parts of our network at a time depending on the available supervisory information within a training batch. See Figure~\ref{fig:dataset} to see sample images from each dataset.

GazeFollow~\cite{nips15_recasens} is a real-world image dataset with manual annotations of the locations where people are looking. The images are taken from other major datasets such as MS COCO~\cite{lin2014microsoft} and PASCAL~\cite{everingham2010pascal}. As a result, the images cover a wide range of scenes, people, and gaze directions. However, the actual 3D gaze angles are not available. Furthermore, images where subjects are looking outside of the image frame are not distinguished and all images have a fixation annotation inside of the frame. Although it is mentioned in \cite{nips15_recasens} that if the annotators indicated that the person was looking outside the image, the image would be discarded, we notice that there are a considerable number of images in which persons appear to be looking outside of the frame. Therefore, we added additional annotations to this dataset in the form of a binary indicator label for ``looking inside'' or ``looking outside'' for every image. In total, we identified 14,564 images correspond to the ``looking outside'' case which is approximately 11.6\% of the total training samples. We have publicly released our additional annotations along with this paper. 

EYEDIAP~\cite{FunesMora_ETRA_2014} dataset is designed for the evaluation of the gaze estimation task. It has videos of 16 different subjects with full face and background visible in a laboratory environment. Each subject was asked to look at a specific target point on a monitor screen and the 3D gaze angle was annotated by leveraging camera calibration and face depth measurement from depth camera. This dataset contains precise 3D gaze angles for frames where the person is fixating the target point. The dataset also contains video of the subjects looking at a 3D ball target instead of 2D screen target point, but we exclude these ball sessions from our experiments in order to conduct a fairer comparison with prior work. We randomly hold out four subjects for test and use the rest of the sessions for training. Since subjects were looking at a screen, all of the frames can be considered as looking outside the image. However, since the dataset has been collected in a controlled setting the backgrounds are primarily white and there is not a lot of variety in lighting or pose. Also, measured gaze angles range between $-40\degree$ to $40\degree$ which is rather limited.

NVIDIA SynHead~\cite{Gu_2017_CVPR} is a synthetic dataset created for the head pose estimation task. The dataset contains 510,960 frames of 70 head motion tracks rendered using 10 individual head models. The gaze of the head is fixed and aligned with the head pose, thus we use the labeled 3D head pose as the gaze angle ground truth. One of the advantages of a synthetic dataset is the ability to insert different images in the background. We randomly generated 15\% from the total frames augmented with provided natural scene backgrounds and regard all as ``looking outside'' examples. The main reason we include SynHead in training is because it complements the EYEDIAP dataset, as the angle ranges are larger, between $-90\degree$ and $90\degree$, and it can include more diverse backgrounds. Since head pose estimation is not a focus of this paper we do not set aside a test set and use SynHead entirely for training. Dataset details are also summarized in Table~\ref{tab:dataset}.

\textit{Training Procedure.} Since each dataset is relevant only to certain subtasks, we only update the relevant parts of the network based on which dataset the training sample is from, while freezing other irrelevant layers during back-propagation. Specifically, when learning gaze angle estimation, we only update the angle pathway (b) and (d) in Figure~\ref{fig:overview}, when learning saliency we update the scene pathway (a), (b) and (c) while freezing all other layers. Similarly, when training fixation likelihood we only update the layer (e) in Figure~\ref{fig:overview}. We found that this selective back-propagation scheme is critical in achieving good performance. 

In every batch, we draw random samples from all of the datasets shuffled together and perform three separate back-propagation for the three outputs as just described. In the beginning, both convolutional pathways were initialized using a ResNet50 model pre-trained on the ImageNet classification task~\cite{deng2009imagenet}. We use the Adam optimization algorithm with a learning rate of $2.5e-4$ and a batch size of 36. Training usually converges within 12 epochs. All of our implementation and experiments are done in PyTorch~\cite{pytorch}.

\setlength{\tabcolsep}{4pt}
\begin{table}[t]
\begin{center}
\caption{Datasets used in our experiments and the number of samples in the training and testing split, as well as the percentage of each split containing peopole looking in/out.}
\label{tab:dataset}
\bgroup
\def\arraystretch{1.25}
\begin{tabular}{l | rr | rr}
\hline
Dataset        & Training set &   & Test set & \\
        &   & in vs out &  & in vs out\\
\hline
GazeFollow~\cite{nips15_recasens}     & 125,557   & 88.4\% vs 11.6\% & 4,782  & 100\% vs 0\% \\
EYEDIAP~\cite{FunesMora_ETRA_2014}        & 72,613     & 0\% vs 100\% & 18,153    & 0\% vs 100\%\\
SynHead~\cite{Gu_2017_CVPR} & 75,400     &  0\% vs 100\%  &  -    & -\\
MMDB~\cite{rehg2013decoding}           & -     & -   & 4,965   &   41.4\% vs 58.6\%\\ \hline
\end{tabular}
\egroup
\end{center}
\end{table}
\setlength{\tabcolsep}{1.4pt}

\section{Evaluation}
In this section we evaluate our model by comparing each output with a number of existing methods and baselines. We first evaluate the person-dependent saliency map in~\ref{sec:gazefollow_eval}, gaze angle estimation in~\ref{sec:eyediap_eval} and general attention estimation in~\ref{sec:mmdb_eval}. Lastly, we evaluate our method by changing model architectures and training dataset in~\ref{sec:model_diagnostic}.

\subsection{Person-Dependent Saliency Prediction}\label{sec:gazefollow_eval}
\begin{figure}[t]
  \centering
      \includegraphics[width=1\textwidth]{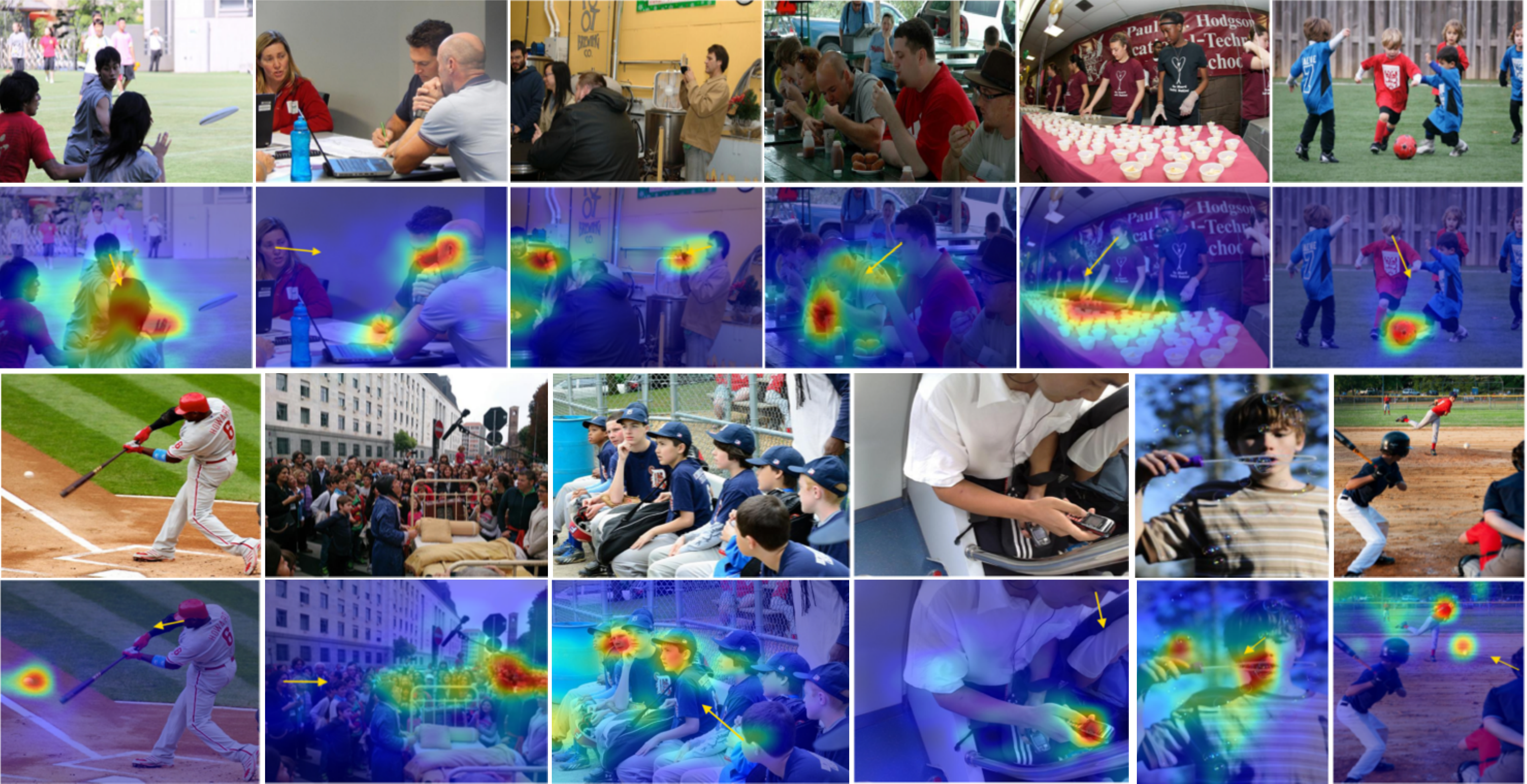}
  \caption{Qualitative results of our model's gaze-saliency prediction on the GazeFollow dataset. Input image is given on the 1st and 3rd row, the output heatmap and estimated gaze is overlaid below.}\label{fig:gazefollow_qual}
\end{figure}

We evaluate the performance of saliency map estimation using the suggested test split of the GazeFollow dataset. The test split contains all ``looking inside'' cases and each test image has multiple gaze target annotations. Following the same evaluation method by~\cite{nips15_recasens}, we compute the Area Under Curve (AUC) score of the Receiver Operating Characteristic (ROC) curve in which the ground truth target positions are the true labels and heatmap value on corresponding positions are prediction confidence score. Our method achieves a score of $0.896$ achieving state-of-the-art performance. Along with AUC we also report results in L2, min distance and angle metric. Please refer to ~\cite{nips15_recasens} for details about the metric. The numbers are summarized in Table~\ref{tab:eval_saliency} along with a number of baselines reported in~\cite{nips15_recasens}. Qualitative results are presented in Figure~\ref{fig:gazefollow_qual}. 

\setlength{\tabcolsep}{4pt}

\begin{table}[t]
\begin{center}
\caption{Gaze-saliency evaluation on the GazeFollow test set}
\label{tab:eval_saliency}
\begin{tabular}{lccc}
\hline\noalign{\smallskip}
Method        & AUC & L2 Distance & Min Distance\\
\noalign{\smallskip}
\hline \noalign{\smallskip}
Random  & 0.504  & 0.484 & 0.391 \\ 
Center  & 0.633  & 0.313 & 0.230 \\
Judd~\cite{judd2009learning}  & 0.711  & 0.337 & 0.250 \\
GazeFollow~\cite{nips15_recasens}   & 0.878  & 0.190 & 0.113 \\
Our  & \textit{0.896}  & \textit{0.187} & \textit{0.112} \\ \hline
\end{tabular}
\end{center}
\end{table}
\setlength{\tabcolsep}{1.4pt}

\subsection{Gaze Angle Prediction}\label{sec:eyediap_eval}
We report the 3D gaze estimation accuracy based on the yaw and pitch output of our model on the chosen EYEDIAP test split. Table~\ref{tab:eval_gaze} shows the angular errors in which we achieve less than 0.5 degrees of difference to the state-of-the-art appearance-based gaze estimation method. It is worth noticing that the middle two values come from~\cite{zhang2017s} which are computed by five-fold cross validation with the entire EYEDIAP dataset whereas our method is evaluated on a single train/test split. Although we did not choose to perform full cross validation, we conclude that it reaches reasonable accuracy on the benchmark. Note also that our method is trained on multiple tasks whereas all other methods are trained solely on the gaze angle prediction task.

\setlength{\tabcolsep}{4pt}
\begin{table}[t]
\begin{center}
\caption{Gaze angle evaluation on EYEDIAP}
\label{tab:eval_gaze}
\begin{tabular}{lc}
\hline\noalign{\smallskip}
Method        & Angular Error (degree)\\
\noalign{\smallskip}
\hline \noalign{\smallskip}
Wood~\cite{wood2015rendering}  & 11.3\degree \\ 
iTracker~\cite{cvpr2016_gazecapture}  & 8.3\degree \\ 
Zhang~\cite{zhang2017s}  & \textit{6.0}\degree \\ 
Our  & 6.4\degree \\ \hline
\end{tabular}

\end{center}
\end{table}
\setlength{\tabcolsep}{1.4pt}

\subsection{Generalized Attention Prediction During Naturalistic Social Interactions}\label{sec:mmdb_eval}
The primary inspiration for our work stems from the need for the ability to quantify various types of visual attention behavior, which is one of the most important nonverbal social cues used in our daily life. Moreover, this is of particular interest among researchers who study child development since gaze behavior of young children is closely related to their social development and developmental disorders such as Autism~\cite{hutman2012selective}. 
The MMDB dataset is one of the largest datasets that contains children's social and communicative behaviors, collected in order to facilitate data-driven analysis of child behavior based on video. The dataset contains a wide range of nonverbal behavior such as hand gestures, smile, and gaze. It has frame-level human annotations of each behavior. As for gaze, each frame is annotated when the child is looking at a ball, book or the examiner. This is done by human annotation based on multiple views, therefore the child's gaze target can be visible or not depending on the viewpoint. Since the annotation does not indicate in which view the gaze target is visible, we added additional annotation ourselves and identified if the target is visible in a child-facing camera view to construct labels for the general attention estimation problem. We publicly release this annotation text file along with our paper. 

We evaluate our method on the generalized attention prediction task. We design a gaze target grid classification task, where each test image is divided into N$\times$N grids. If the subject is looking inside of the image then the grid square which contains the gaze target is assigned a label of 1 while others are assigned labels of 0. If the subject is looking somewhere outside of the frame then all grid squares are assigned the 0 label. Using our method's fixation likelihood map we predict the positive gaze grid square. We test the GazeFollow model~\cite{nips15_recasens} which is the closest work to our method in terms of having the ability to predict gaze target location. One of its limitations is the inability to correctly predict the ``outside'' case, where the subject is looking outside of the frame. As a result, our method achieves much higher precision in addition to increased recall as shown in Table~\ref{tab:eval_mmdb_grid}.

Additionally, we constructed various baseline tests consisting of a classifier based on a subset of features constructed for saliency, gaze and head pose. Specifically, we tested with SVM and Random Forest using a subset of \{\cite{nips15_recasens},~\cite{zhang2017s},~\cite{baltruvsaitis2016openface}\} as features. In other words, each classifier has been trained for detection of looking inside with the training set described in Table~\ref{tab:dataset}, using aforementioned features, and tested on the MMDB images. 
We report the results in Table~\ref{tab:eval_mmdb_likelihood}. Note that the MMDB dataset was not used for training across all methods including ours. 
Moreover, we evaluate the value of multi-dataset training in solving the general attention problem. As shown in the last three rows of Table~\ref{tab:eval_mmdb_likelihood}, joint training of gaze and saliency is critical in solving the general attention estimation task because without the gaze angle estimate it is ineffective to determine whether the subject is looking inside or outside the frame.

\setlength{\tabcolsep}{4pt}
\begin{table}[t]
\begin{center}
\caption{Evaluation on MMDB - gaze target grid classification}
\label{tab:eval_mmdb_grid}
\begin{tabular}{ llccc}
\hline\noalign{\smallskip}
Grid Size & Method        & Precision & Recall\\
\noalign{\smallskip}
\hline \noalign{\smallskip}
\hline
2x2 & GazeFollow~\cite{nips15_recasens} & 0.344  & 0.715\\ 
 & Our  & 0.744  & 0.851\\
\noalign{\smallskip}
\hline \noalign{\smallskip}
\hline
 5x5 & GazeFollow~\cite{nips15_recasens} & 0.210  & 0.437\\ 
 & Our  & 0.614  & 0.683\\ \hline
\end{tabular}
\end{center}
\end{table}
\setlength{\tabcolsep}{1.4pt}

\begin{figure}[t]
  \centering
      \includegraphics[width=1\textwidth]{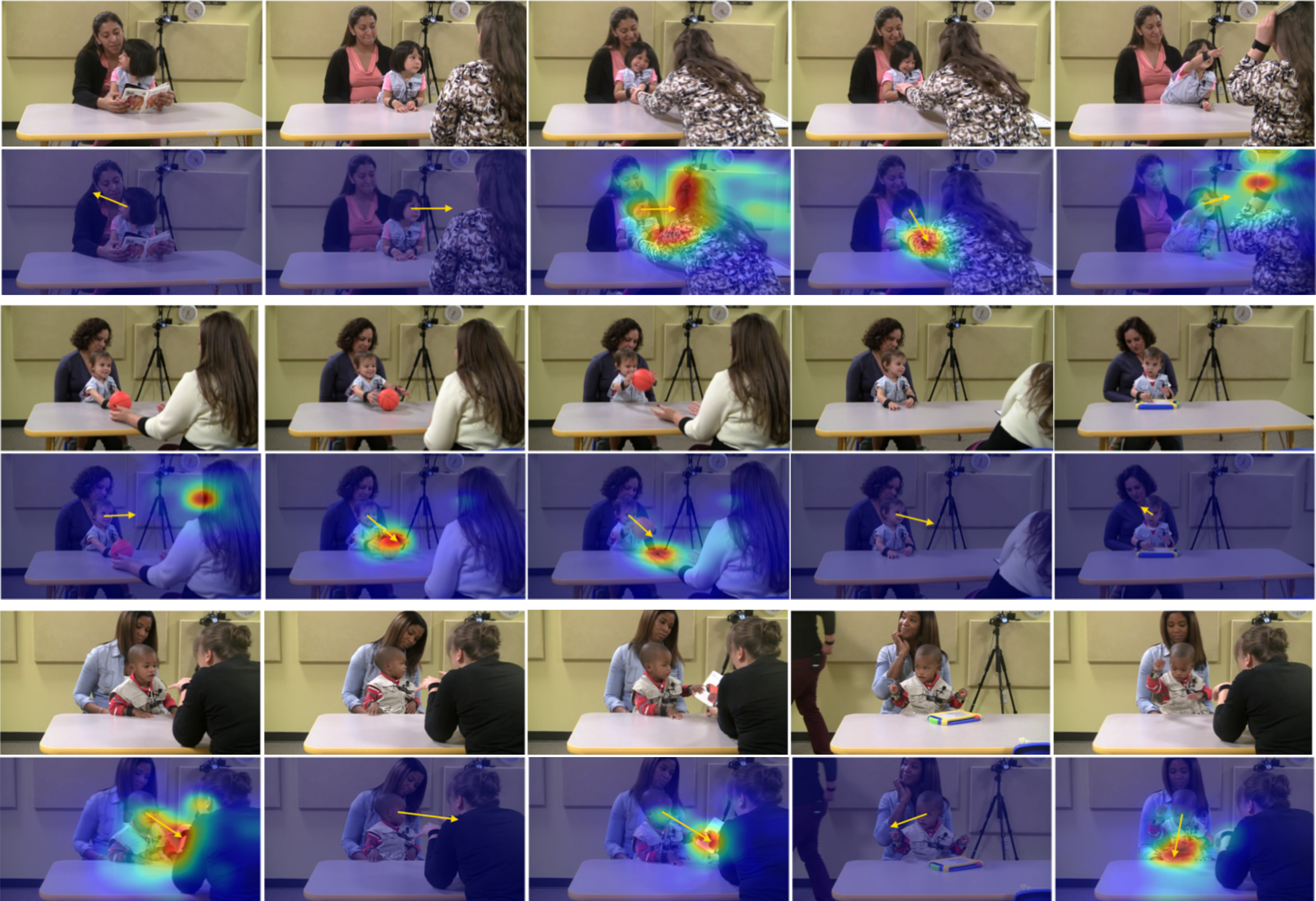}
  \caption{Example result of our method on the MMDB dataset. The dataset contains various types of gaze behavior including fixations on a target both within and out of frame. Our method produces low heatmap when the fixation target is outside and high heatmap when the target becomes clear.}\label{fig:mmdb_good}
\end{figure}

\setlength{\tabcolsep}{4pt}
\begin{table}
\begin{center}
\caption{Evaluation of fixation likelihood on MMDB}
\label{tab:eval_mmdb_likelihood}
\begin{tabular}{llcc}
\hline \noalign{\smallskip}
Method        &   & Average Precision \\
\noalign{\smallskip}
\hline \noalign{\smallskip}
SVM with GazeFollow~\cite{nips15_recasens} &  & 0.311    \\ 
SVM with GazeFollow~\cite{nips15_recasens}+gaze~\cite{zhang2017s} &  & 0.531   \\ 
SVM with GazeFollow~\cite{nips15_recasens}+headpose~\cite{baltruvsaitis2016openface} &  & 0.620    \\
SVM with gaze~\cite{zhang2017s}+headpose~\cite{baltruvsaitis2016openface} &  &   0.405  \\
SVM with GazeFollow~\cite{nips15_recasens}+gaze~\cite{zhang2017s}+headpose~\cite{baltruvsaitis2016openface} &  & 0.624      \\ \hline
Random Forest with GazeFollow~\cite{nips15_recasens} &  & 0.707    \\
Random Forest with GazeFollow~\cite{nips15_recasens}+gaze~\cite{zhang2017s} &  & 0.727    \\
Random Forest with GazeFollow~\cite{nips15_recasens}+headpose~\cite{baltruvsaitis2016openface} &  & 0.785    \\
Random Forest with gaze~\cite{zhang2017s}+headpose~\cite{baltruvsaitis2016openface} &  &    0.512 \\
Random Forest with GazeFollow~\cite{nips15_recasens}+gaze~\cite{zhang2017s}+headpose~\cite{baltruvsaitis2016openface} &  & 0.773      \\ \hline
Our, trained only with GazeFollow dataset                &  & 0.737      \\
Our, trained only with GazeFollow and EYEDIAP dataset                &  & 0.820      \\
Our final               &  & \textit{0.902}      \\ \hline
\end{tabular}
\end{center}
\end{table}
\setlength{\tabcolsep}{1.4pt}

\subsection{Alternative Model and Diagnostics}\label{sec:model_diagnostic}
Finally, we run additional experiments to study how the performance of our model is affected by different training datasets and architectural choice by evaluating it on the GazeFollow benchmark. As shown in Table~\ref{tab:eval_more}, omitting EYEDIAP or SynHead training dataset did not have much impact on the attention-within-an-image heatmap estimation whereas changing model architecture considerably affected the scores. For example, using a single ResNet50 pathway which pools facial features using ROI-pooling shows significantly degraded performance which supports our decision to use a scene pathway as well as a face pathway. 
Interestingly, the project-and-compare loss was not as helpful as initially expected, and we think that this is because the coverage range of pose in the SynHead and EYEDIAP datasets is limited (within $\pm~90$) which is not the case in the GazeFollow dataset.

\begin{figure}[h]
  \centering
      \includegraphics[width=1\textwidth]{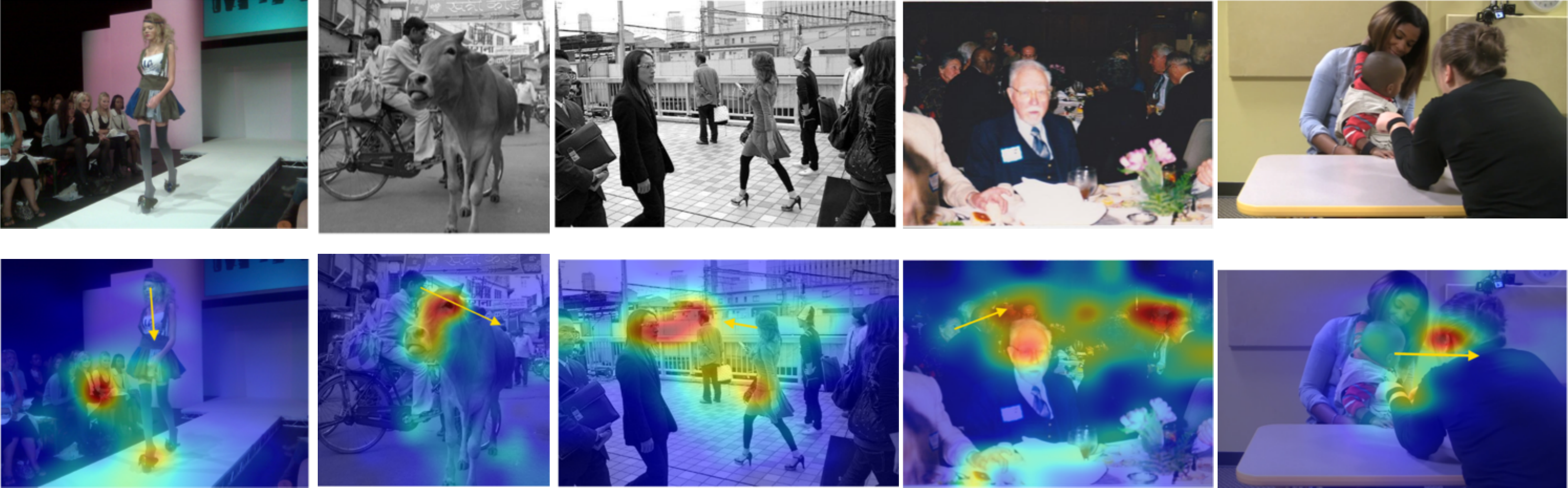}
  \caption{Challenging cases due to occlusion and the lack of depth understanding.}\label{fig:failcases}
\end{figure}

Qualitatively, we were able to observe that, even though our method is designed to measure fixation outside, it can make mistakes when the target is within the frame but occluded by other object. Also, when the subject is closer to the camera than some salient object in the background, the method sometimes estimates those as fixation candidate due to the lack of scene depth understanding. Examples are illustrated in Figure~\ref{fig:failcases}.

\setlength{\tabcolsep}{4pt}
\begin{table}[t]
\begin{center}
\caption{Additional model evaluation and diagnostics on the GazeFollow test split}
\label{tab:eval_more}
\bgroup
\def\arraystretch{1}
\begin{tabular}{l  cc }
\noalign{\smallskip}
\hline \noalign{\smallskip}
Method        & AUC & L2 Distance \\
\noalign{\smallskip}
\hline \noalign{\smallskip}
No EYEDIAP & 0.887  & 0.197 \\
No SynHead & 0.895  & 0.191\\
No EYEDIAP and SynHead & 0.891  & 0.194 \\
No project-and-compare loss & 0.895  & 0.189 \\
Map resolution 15x15 & 0.778  & 0.194\\
ROI-pooling & 0.700  & 0.325\\
\hline \noalign{\smallskip}
Our final & \textit{0.896}  & \textit{0.187} \\
\hline \noalign{\smallskip}
\end{tabular}
\egroup
\end{center}
\end{table}
\setlength{\tabcolsep}{1.4pt}

\section{Conclusion}
In this paper we presented the new challenging problem of generalized visual attention prediction which encapsulates several constrained attention prediction and gaze estimation problems that have been the focus of many previous works. We proposed a multi-task learning approach and neural architecture leveraging three different datasets which tackles this problem and works across multiple naturalistic social scenarios. In order to train our architecture we have supplemented these datasets with new annotations and we release these annotations to the public. Our model achieves state-of-the-art performance on the single-task gaze-saliency prediction and competes with state-of-the-art methods on gaze estimation benchmarks while achieving promising performance on the generalized attention prediction problem on the MMDB dataset. Future work in this area can lead to breakthroughs in attention prediction applications which are valuable in numerous scientific and commercial areas. A suggested first step would be to improve existing datasets with additional annotations or collect datasets tailored for this problem.

\section{Acknowledgement}
This study was funded in part by the Simons Foundation under grant 247332.

%
%
%
\bibliographystyle{splncs04}
\bibliography{mybibliography}

\begin{thebibliography}{10}
\providecommand{\url}[1]{\texttt{#1}}
\providecommand{\urlprefix}{URL }
\providecommand{\doi}[1]{https://doi.org/#1}

\bibitem{pytorch}
Pytorch: Tensors and dynamic neural networks in python with strong gpu
  acceleration. \url{https://github.com/pytorch/pytorch}, accessed: 2017-11-03

\bibitem{baltruvsaitis2016openface}
Baltru{\v{s}}aitis, T., Robinson, P., Morency, L.P.: Openface: an open source
  facial behavior analysis toolkit. In: Applications of Computer Vision (WACV),
  2016 IEEE Winter Conference on. pp. 1--10. IEEE (2016)

\bibitem{benfold2009attention}
Benfold, B., Reid, I.: Guiding visual surveillance by tracking human attention.
  In: British Machine Vision Conference (September 2009)

\bibitem{borji2015salient}
Borji, A., Cheng, M.M., Jiang, H., Li, J.: Salient object detection: A
  benchmark. IEEE Transactions on Image Processing  \textbf{24}(12),
  5706--5722 (2015)

\bibitem{borji2013stands}
Borji, A., Sihite, D.N., Itti, L.: What stands out in a scene? a study of human
  explicit saliency judgment. Vision research  \textbf{91},  62--77 (2013)

\bibitem{chen2016subjects}
Chen, C.Y., Grauman, K.: Subjects and their objects: Localizing interactees for
  a person-centric view of importance. International Journal of Computer Vision
  pp. 1--22 (2016)

\bibitem{chong2017detecting}
Chong, E., Chanda, K., Ye, Z., Southerland, A., Ruiz, N., Jones, R.M., Rozga,
  A., Rehg, J.M.: Detecting gaze towards eyes in natural social interactions
  and its use in child assessment. Proceedings of the ACM on Interactive,
  Mobile, Wearable and Ubiquitous Technologies  \textbf{1}(3), ~43 (2017)

\bibitem{Cristani}
Cristani, M., Bazzani, L., Paggetti, G., Fossati, A., Tosato, D., Del~Bue, A.,
  Menegaz, G., Murino, V.: Social interaction discovery by statistical analysis
  of f-formations. In: Proc. BMVC (2011)

\bibitem{deng2009imagenet}
Deng, J., Dong, W., Socher, R., Li, L.J., Li, K., Fei-Fei, L.: Imagenet: A
  large-scale hierarchical image database. In: Computer Vision and Pattern
  Recognition, 2009. CVPR 2009. IEEE Conference on. pp. 248--255. IEEE (2009)

\bibitem{everingham2010pascal}
Everingham, M., Van~Gool, L., Williams, C.K., Winn, J., Zisserman, A.: The
  pascal visual object classes (voc) challenge. International journal of
  computer vision  \textbf{88}(2),  303--338 (2010)

\bibitem{FunesMora_ETRA_2014}
Funes~Mora, K.A., Monay, F., Odobez, J.M.: Eyediap: A database for the
  development and evaluation of gaze estimation algorithms from rgb and rgb-d
  cameras. In: Proceedings of the ACM Symposium on Eye Tracking Research and
  Applications. ACM (Mar 2014). \doi{10.1145/2578153.2578190}

\bibitem{gorji2017attentional}
Gorji, S., Clark, J.J.: Attentional push: A deep convolutional network for
  augmenting image salience with shared attention modeling in social scenes.
  In: Proceedings of the IEEE Conference on Computer Vision and Pattern
  Recognition. pp. 2510--2519 (2017)

\bibitem{Gu_2017_CVPR}
Gu, J., Yang, X., De~Mello, S., Kautz, J.: Dynamic facial analysis: From
  bayesian filtering to recurrent neural network. In: The IEEE Conference on
  Computer Vision and Pattern Recognition (CVPR) (July 2017)

\bibitem{He2015}
He, K., Zhang, X., Ren, S., Sun, J.: Deep residual learning for image
  recognition. arXiv preprint arXiv:1512.03385  (2015)

\bibitem{hutman2012selective}
Hutman, T., Chela, M.K., Gillespie-Lynch, K., Sigman, M.: Selective visual
  attention at twelve months: Signs of autism in early social interactions.
  Journal of autism and developmental disorders  \textbf{42}(4),  487--498
  (2012)

\bibitem{itti1998model}
Itti, L., Koch, C., Niebur, E.: A model of saliency-based visual attention for
  rapid scene analysis. IEEE Transactions on pattern analysis and machine
  intelligence  \textbf{20}(11),  1254--1259 (1998)

\bibitem{judd2009learning}
Judd, T., Ehinger, K., Durand, F., Torralba, A.: Learning to predict where
  humans look. In: Computer Vision, 2009 IEEE 12th international conference on.
  pp. 2106--2113. IEEE (2009)

\bibitem{cvpr2016_gazecapture}
Krafka, K., Khosla, A., Kellnhofer, P., Kannan, H., Bhandarkar, S., Matusik,
  W., Torralba, A.: Eye tracking for everyone. In: IEEE Conference on Computer
  Vision and Pattern Recognition (CVPR) (2016)

\bibitem{land2009looking}
Land, M., Tatler, B.: Looking and acting: vision and eye movements in natural
  behaviour. Oxford University Press (2009)

\bibitem{li2015visual}
Li, G., Yu, Y.: Visual saliency based on multiscale deep features. In:
  Conference on Computer Vision and Pattern Recognition (2015)

\bibitem{li2014secrets}
Li, Y., Hou, X., Koch, C., Rehg, J.M., Yuille, A.L.: The secrets of salient
  object segmentation. In: Proceedings of the IEEE Conference on Computer
  Vision and Pattern Recognition. pp. 280--287 (2014)

\bibitem{lin2014microsoft}
Lin, T.Y., Maire, M., Belongie, S., Hays, J., Perona, P., Ramanan, D.,
  Doll{\'a}r, P., Zitnick, C.L.: Microsoft coco: Common objects in context. In:
  European conference on computer vision. pp. 740--755. Springer (2014)

\bibitem{nips15_recasens}
Recasens$^*$, A., Khosla$^*$, A., Vondrick, C., Torralba, A.: Where are they
  looking? In: Advances in Neural Information Processing Systems (NIPS) (2015),
  $^*$ indicates equal contribution

\bibitem{Recasens_2017_ICCV}
Recasens, A., Vondrick, C., Khosla, A., Torralba, A.: Following gaze in video.
  In: The IEEE International Conference on Computer Vision (ICCV) (Oct 2017)

\bibitem{rehg2013decoding}
Rehg, J., Abowd, G., Rozga, A., Romero, M., Clements, M., Sclaroff, S., Essa,
  I., Ousley, O., Li, Y., Kim, C., et~al.: Decoding children's social behavior.
  In: Proceedings of the IEEE conference on computer vision and pattern
  recognition. pp. 3414--3421 (2013)

\bibitem{soo2015social}
Soo~Park, H., Shi, J.: Social saliency prediction. In: Proceedings of the IEEE
  Conference on Computer Vision and Pattern Recognition. pp. 4777--4785 (2015)

\bibitem{sugano2014learning}
Sugano, Y., Matsushita, Y., Sato, Y.: Learning-by-synthesis for
  appearance-based 3d gaze estimation. In: Proceedings of the IEEE Conference
  on Computer Vision and Pattern Recognition. pp. 1821--1828 (2014)

\bibitem{wang2015deep}
Wang, L., Lu, H., Ruan, X., Yang, M.H.: Deep networks for saliency detection
  via local estimation and global search. In: Computer Vision and Pattern
  Recognition (CVPR), 2015 IEEE Conference on. pp. 3183--3192. IEEE (2015)

\bibitem{wood2015rendering}
Wood, E., Baltrusaitis, T., Zhang, X., Sugano, Y., Robinson, P., Bulling, A.:
  Rendering of eyes for eye-shape registration and gaze estimation. In:
  Proceedings of the IEEE International Conference on Computer Vision. pp.
  3756--3764 (2015)

\bibitem{zhang2017everyday}
Zhang, X., Sugano, Y., Bulling, A.: Everyday eye contact detection using
  unsupervised gaze target discovery. In: 30th Annual Symposium on User
  Interface Software and Technology. ACM (2017)

\bibitem{zhang15_cvpr}
Zhang, X., Sugano, Y., Fritz, M., Bulling, A.: Appearance-based gaze estimation
  in the wild. In: Proc. of the IEEE Conference on Computer Vision and Pattern
  Recognition (CVPR). pp. 4511--4520 (June 2015)

\bibitem{zhang2017s}
Zhang, X., Sugano, Y., Fritz, M., Bulling, A.: It's written all over your face:
  Full-face appearance-based gaze estimation. In: Proc. IEEE International
  Conference on Computer Vision and Pattern Recognition Workshops (CVPRW)
  (2017)

\bibitem{zhao2015saliency}
Zhao, R., Ouyang, W., Li, H., Wang, X.: Saliency detection by multi-context
  deep learning. In: Proceedings of the IEEE Conference on Computer Vision and
  Pattern Recognition. pp. 1265--1274 (2015)

\end{thebibliography}

\end{document}